\documentclass[11pt,letterpaper]{article}
\usepackage[T1]{fontenc}
\usepackage{lmodern}
\usepackage[utf8]{inputenc}
\usepackage[margin=1in]{geometry}
\usepackage{amsmath}
\usepackage{booktabs}
\usepackage{graphicx}
\usepackage{tabularx}
\usepackage{enumitem}
\usepackage[hidelinks]{hyperref}
\usepackage{natbib}
\setcitestyle{numbers,square}
\setlist{itemsep=1pt,topsep=3pt}

\title{Aerial Person Detection under Edge Constraints:\\Held-Out Evaluation and Edge--Ground Verification}
\author{Raghu Venkat\thanks{Numberz.ai Inc., Huntsville, AL, USA. \texttt{raghu.venkat@numberz.ai}}
\and Tricha Anjali\thanks{Numberz.ai Inc., Huntsville, AL, USA. \texttt{tricha.anjali@numberz.ai}}}
\date{November 2026}

\begin{document}
\maketitle

\begin{abstract}
A search drone sees a missing person as a few pixels, carries a small computer, and shares a narrow radio link. We study person detection under those limits; the primary evaluations were declared before they ran, and later analyses are marked exploratory. In an exploratory comparison, holding out recording blocks inferred from thermal frame identifiers cut recall at a fixed false-alarm target from 32.7\% to 12.3\%. With nothing tuned on test blocks, a 46,000-parameter detector gated by a contrast cue finds 10.4\% of people against 5.7\% for the same network applied everywhere. A standard detector, YOLO11n, trained under the same folds finds 54.5\% at a similar false-alarm rate for about eleven times the arithmetic, and on our 4K field footage at 117~m has a walking person among its top 16 detections in up to 92\% of frames against 74\% for our board. Run alone, YOLO11n's thresholds set at one location overshot their false-alarm target 1.7 to 3.4 times at new locations, and its best configuration at altitude cost 4 to 24 standard passes per frame. Re-scoring its 16 best crops (154~KB per frame) with an 842,000-parameter ground model, thresholds frozen on a validation location, raised recall from 44.9\% to 53.7\% against the same 16 crops with YOLO's own scores; in a post hoc paired analysis over 15 test locations the gain was +8.8 points [+2.4, +12.5]. Pooled false detections fell from 0.86 to 0.64 per frame, but that paired change was not distinguishable from zero. Coverage of a still person in earth tones was very low for both detectors in the clips we flew.
\end{abstract}

\section{Introduction}
When a hiker goes missing, a search team needs to cover ground faster than people on foot can. Small drones help, but each one sends live video to a pilot who watches one screen. Three limits decide what a small drone can do on its own. A person seen from 100~m or more covers a few pixels. The computer on board is small and draws from the flight battery. The radio carries a fraction of one video stream, and less when several drones share it.

Three questions follow. How well does a detector that fits on a Pi-class board find people in recordings and places it was not tuned on, at a false-alarm rate an operator can live with? How does a standard detector do under the same conditions, and what does it cost? And when the board falls short, how much does sending a few small crops to a larger model on the ground recover, and at what cost in bytes?

We answer them on public data and on our own footage. The primary evaluations were written down with their read-outs before they ran; analyses added later are labelled exploratory. The contributions fall under three themes.
\begin{enumerate}
\item \emph{Generalization and evaluation integrity.} We check two public aerial datasets for overlap between their published training and test splits, evaluate on folds that hold out inferred recording blocks (thermal) and whole locations (mountain search imagery), and measure how far thresholds set at one location drift at others.
\item \emph{Accuracy and resource trade-offs.} We compare a cue-gated 46,000-parameter detector with the same network applied everywhere, train a standard detector (YOLO11n~\cite{yolo11}) under the same folds, run both on 4K field footage at four heights, and report measured board power for our pipeline.
\item \emph{Selective ground verification.} We measure a ground model that re-scores the board's 16 best crops, behind our detector and behind YOLO11n, with frozen thresholds, a same-crops control and a shuffled-label control; we report its benefit, its transmission cost and what remains to be validated.
\end{enumerate}
This is a measurement study. It does not propose a new detector architecture, and the edge--ground split it measures is an established pattern (Section~\ref{sec:related}). It does not validate an autonomous search system. For people hidden under forest canopy, airborne optical sectioning~\cite{schedl2021} is the stronger approach and the one we plan to build on; our study covers fast, wide-area colour search, where it does not apply. Our own board detector is beaten by the standard one on accuracy; we report that plainly. What holds up is the ground verification step behind YOLO11n: over the same candidates it raised recall, with the paired change across locations clear of zero. Its effect on false detections was mixed across locations. The study builds on our earlier work on gated sensing on airborne video~\cite{venkat2026sisr}, which reported where such pipelines lose evidence.

\section{Related work}\label{sec:related}
\paragraph{People from the air.} Bo\v{z}i\'c-\v{S}tuli\'c et al.\ introduced HERIDAL, aerial images from mountain search exercises, and showed that deep detectors find people in it~\cite{heridal}. Sambolek and Ivasic-Kos built SARD for the same purpose~\cite{sard}. Broyles et al.\ released WiSARD, paired visual and thermal images from wilderness search flights~\cite{wisard}. Suo et al.\ released HIT-UAV, a high-altitude thermal set~\cite{hituav}, and Makrigiorgis et al.\ a set of small people seen from a drone~\cite{kios}. Manzini and Murphy set out open problems for computer vision in wilderness search, drawn from an actual search~\cite{manzini2023open}, and Murphy and Manzini described how drone imagery should be collected for such work~\cite{murphy2023improving}. Yu et al.\ studied tiny people in wide images~\cite{tinyperson}.

\paragraph{Hidden people and real searches.} Three works bear directly on our findings (Table~\ref{tab:related}). Bimber's group developed airborne optical sectioning (AOS), in which thermal images taken along a flight path are combined into a synthetic aperture that suppresses occluding foliage, and Schedl, Kurmi and Bimber applied it to search and rescue~\cite{schedl2020}. Their autonomous drone ran the integration, a YOLOv4-tiny classifier and route adaptation on a Raspberry~Pi~4 with a neural compute stick, sent only classification results over the link, and found 38 of 42 hidden people in 17 field experiments with one false alert on predefined paths~\cite{schedl2021}. It is, to our knowledge, the strongest result for people under forest canopy, and an earlier demonstration of on-board detection with a minimal link. Its demonstrated envelope differs from wide-area search: flights at about 35~m and 1~m/s along predefined paths, about 6~ha over eight flights, people who stay still while the aperture is integrated, a body warmer than its surroundings, and accurate camera poses for every image. How it performs at 100~m and above, in warm conditions, or over larger areas per flight was not part of that study. Manzini and Murphy applied a detector that scored 78.6\% precision and 84.0\% recall on HERIDAL, and an anomaly detector, to 18,759 frames from the search for Patricia Wu-Murad; neither produced a useful lead, and the detector missed members of the search team while labelling bushes, rocks and branches as people~\cite{manzini2023open}. They describe conditions benchmarks rarely contain: imagery collected during an operation, oblique flights below the treeline where canopy was dense, a wide variety of imagery, and no agreed measure of performance. They call for more realistic data, models that handle that variety, and agreement on metrics. Nathan et al.\ released colour imagery from a police search at about 300~m over forest, on which anomaly detectors reached precision below 3.5\% and a supervised YOLOv12 model failed~\cite{nathan2026anomaly}.

\begin{table}[h]\centering\small
\caption{Closest prior work, as reported by its authors. Settings differ, so this is not a head-to-head comparison.}\label{tab:related}
\begin{tabularx}{\linewidth}{@{}>{\raggedright\arraybackslash}p{2.6cm}>{\raggedright\arraybackslash}p{3.4cm}>{\raggedright\arraybackslash}X@{}}\toprule
Work & Setting & Reported result \\\midrule
Schedl et al.~\cite{schedl2021} & Thermal AOS, about 35~m, 1~m/s, Pi~4 on board & 38 of 42 hidden people found; one false alert; about 6~ha over 8 flights \\
Manzini and Murphy~\cite{manzini2023open} & Colour video from a real search, offline & Benchmark-strong detector gave no useful lead in the field \\
Nathan et al.~\cite{nathan2026anomaly} & Colour, about 300~m, 4~cm/px, offline & Anomaly detectors below 3.5\% precision; YOLOv12 failed \\
This paper & Colour, 9 to 117~m and public sets, Pi~5 & Held-out and frozen-threshold evaluation; ground verification +8.8 points recall over the same candidates; still person in earth tones 0 to 7.5\% coverage \\\bottomrule
\end{tabularx}
\end{table}

Our still-person result agrees with all three: colour detection rarely finds a still or partly hidden person, and benchmark scores do not carry to the field. What we add is a measurement of that gap under held-out locations and frozen thresholds, and of what ground verification buys at a known link cost. We do not test AOS.

\paragraph{Cheap stages ahead of expensive models.} Viola and Jones made the attentional cascade standard in detection~\cite{violajones}. Kang et al.\ placed cheap models ahead of an expensive detector to skip video frames on a server~\cite{kang2017noscope}. Event-triggered control trades how often a system acts against how well it performs~\cite{heemels2012}. Splitting a network or a pipeline between a device and a server is studied as split computing or edge--cloud inference~\cite{kang2017neurosurgeon,matsubara2022split}; our edge--ground split is an instance of it, and a second stage that re-scores a first stage's candidates is the classic cascade. Single-stage detectors of the YOLO family are the usual choice for small boards~\cite{yolo11}; we use one as the baseline.

\paragraph{Evaluation.} Kaufman et al.\ describe leakage between training and test data and how it inflates results~\cite{kaufman2012leakage}. Aerial frames taken seconds apart from one flight are near-duplicates, so a frame-level split lets the same scene appear on both sides. Evaluation on shared recordings and on unseen recordings answer different questions; both are legitimate, and we report the second because a search drone flies over ground it has not seen.

\section{Data and splits}
\paragraph{HIT-UAV (thermal).} 2,898 long-wave infrared frames with 12,312 labelled people~\cite{hituav}. We infer recording blocks from the frame identifiers: they run in 39 contiguous blocks, each with one light condition and one altitude, and each block sweeps the camera angles. We treat each block as one recording session. The dataset does not label sessions; this grouping is inferred by us and not validated against flight logs, so held-out blocks are not established to be independent flights or unseen sites. The block list is in the released truth register. All 39 blocks have frames in both the published training and test splits. We use five folds of whole inferred blocks, balanced on labelled people. Within each training pool, one block in four (by a seeded hash) is held back for validation.

\paragraph{HERIDAL (mountain search).} We use the repackaging of HERIDAL by Pyrro (Zenodo record 5662351), which its release page describes as a modified version. In this release, 66 of the 103 published test images are byte-identical to training images, and the test-only location codes are aliases of training codes (GRO of GOR, MOB of BLA). We make no claim about other distributions of HERIDAL. After dropping duplicates and merging aliases we keep 1,582 frames, 3,091 people and 15 locations, in three folds of whole locations. One further training location per fold is held out for validation. The duplicate pairs, the alias map and the split manifest are in the released truth register.

\paragraph{Our footage.} Hunter Klobucar, an FAA Part~107 certificated remote pilot, flew a drone recording 3840-pixel-wide (4K) video at 29.97 frames per second, camera pointing down, over a trail and tree edge. On the first day a person walked the trail at 117.5, 89.2, 60.3 and 9.3~m (195 labelled samples). On the second, one person stayed still in four poses at 117, 90, 60 and 30~m, with and without a bright vest (276 samples). No model saw this site or camera in training. Truth for the walker comes from frame differencing re-centred on the body and checked by eye. Truth for the still person comes from one marked point per clip carried by image registration.

\paragraph{Hits and false alarms.} Every detector outputs scored points; for box detectors the point is the box centre. A point is a hit when it lies inside a labelled box grown by 1.5~px on each side. Matching is one to one and greedy by descending score. A second point on an already matched person counts as a duplicate, not a false alarm. A point inside no grown box and no ignore region is a false alarm. In the field, a hit is a point within 0.25~m on the ground of the labelled position. We report 95\% intervals from a bootstrap over inferred blocks or locations, not frames. Learned arms have a shuffled-label control with the same architecture and data; for the ground model behind YOLO11n the control was trained for the same 12 epochs as the real model (an earlier 3-epoch control is also reported).

\paragraph{Two terms.} \emph{Candidate coverage} is the share of people present among the top 16 detections or crops. \emph{Detection recall} is the share the full system reports under the stated false-alarm rule. Field numbers in this paper are candidate coverage.

\section{Method}
\paragraph{Cue.} Each pixel passes a local-contrast and blob-size test at full and half resolution. A blob is kept when it stands at least $z$ local standard deviations and $\delta$ grey levels above its 41-pixel background, with an area of 2 to 900 pixels. Blobs within 4~px are merged and the strongest 128 are kept (384 on colour imagery, where a colour-strength channel is searched as well). On thermal folds, $z$ and $\delta$ are chosen inside each fold (Section~\ref{sec:boundary}).

\paragraph{Gated model.} A 46,000-parameter convolutional network (five 3$\times$3 convolutions and one 6$\times$6, with batch normalization and two pooling stages, on a 44$\times$44 crop) scores the kept places. In the deployed stream, one frame in ten also gets a pass of the network over the whole frame, so that an area the cue misses is still checked.

\paragraph{Dense pass.} The comparison arm runs the same network fully convolutionally over the whole frame at one scale, with reflected borders, giving a score every 4 pixels: about 20,000 positions on a 640$\times$512 thermal frame.

\paragraph{Standard detector.} YOLO11n (Ultralytics, 2.6~M parameters, 6.4~GFLOPs at 640, COCO-pretrained), single class person, trained per fold on the same training blocks or locations, 100 epochs with patience 20 on the same validation data, defaults otherwise; imgsz 640 on thermal and 800 on HERIDAL. On 4K field frames we run it four ways: the whole frame at 640 (W640) and 1280 (W1280); all 24 tiles of 640$\times$640 at full resolution, merged (T); and 640$\times$640 windows centred on our board's 16 best points (G). We report each as 640-equivalent passes per frame: 1, 4, 24 and about 9 to 11. On the field footage we also run the COCO weights unchanged. Ultralytics is AGPL-3.0; the baseline is used for comparison and is not part of our release.

\paragraph{Edge--ground split.} The board sends its 16 highest-scoring places as 192$\times$192 crops (at the 800$\times$600 working scale) at JPEG quality 85; the ground resizes each to 96$\times$96. A convolutional confirmer of 842,209 parameters (0.43 GMAC per crop), trained from scratch on each fold's training locations, re-scores them. The number of crops was fixed at 16 before any run. Behind YOLO11n, the same recipe is retrained on YOLO's own top points from the training frames.

Training details are in Appendix~\ref{app:train}.

\subsection{What each part was tuned on}\label{sec:boundary}
A paper about leakage needs an exact account of its own development and test boundary (Table~\ref{tab:boundary}).

\begin{table}[h]\centering\small
\caption{Development and test boundary.}\label{tab:boundary}
\begin{tabularx}{\linewidth}{@{}lX@{}}\toprule
Component & Data that informed it \\\midrule
Thermal cue settings & Nine settings fixed in advance; one chosen per fold from that fold's training and validation blocks only. \\
Thermal networks, YOLO11n & Training blocks or locations of each fold. \\
Thresholds & Validation blocks or location of each fold, then frozen. \\
Memory settings & Validation blocks (memory was not re-run under the per-fold cue). \\
Colour cue (HERIDAL, field) & Fixed during work on a different dataset~\cite{kios}, before any HERIDAL frame was used. \\
HERIDAL test-curve read-out & Chosen after the first run; reported as secondary. \\
Field check & Nothing; models, cue and radius fixed before the footage was scored. \\\bottomrule
\end{tabularx}
\end{table}

An earlier run chose the thermal cue on 200 frames that sit in held-out blocks of three folds; its numbers are in the protocol file as exploratory and are not the headline here.

\section{Results}
\subsection{Split overlap}
Table~\ref{tab:leak} scores the same detector on the published HIT-UAV split and on folds that hold out inferred blocks, with thresholds set for 0.1 false alarms per frame on validation. The published split gives about two and a half times the recall. Both rows use the earlier cue, which was chosen on frames that lie in held-out blocks of three folds, so this comparison is exploratory. It shows a large difference between the two evaluation protocols; it is not a clean measure of leakage alone. On HERIDAL the duplicates moved our numbers by under one point, because we had already held out locations.

\begin{table}[h]\centering\small
\caption{HIT-UAV, native resolution, earlier cue (exploratory), thresholds for 0.1 false alarms per frame on validation. Actual test false alarms per frame in brackets.}\label{tab:leak}
\begin{tabular}{@{}lcc@{}}\toprule
Split & Gated & Dense \\\midrule
Published split (blocks shared) & 32.7\% (0.11) & 25.2\% (0.12) \\
Inferred blocks held out, five folds & 12.3\% (0.18) & 6.5\% (0.17) \\
Shuffled-label control, held out & 0.0\% (0.07) & 0.0\% (0.02) \\\bottomrule
\end{tabular}
\end{table}

\subsection{Thermal: gating, and a standard detector}
With the cue chosen inside each fold, the gated detector finds 10.4\% [6.6, 15.1] of people at 0.19 false alarms per frame and the dense network 5.7\% [3.1, 9.6] at 0.17; the paired gain is +4.7 points [+2.4, +6.6] (Table~\ref{tab:gate}). The gated arm uses about a fifth of the dense arm's arithmetic. The cue keeps 90.9\% of people among its candidates, so most misses happen in the network, not the cue. Short-term memory across frames, measured under the earlier cue, lowered false alarms from 0.20 to 0.11 per frame (interval $-0.15$ to $-0.05$); its recall gain was not clear of zero.

YOLO11n trained under the same folds finds 54.5\% [49.9, 58.9] at 0.14 false alarms per frame, about five times our gated detector, for about 3.2 GMAC per frame against 0.28. On thermal imagery our board detector is not competitive with a standard detector on accuracy.

\begin{table}[h]\centering\small
\caption{HIT-UAV, inferred recording blocks held out, five folds pooled, native resolution, thresholds for 0.1 false alarms per frame on validation blocks. Arithmetic is network multiply-adds per frame.}\label{tab:gate}
\begin{tabular}{@{}lccc@{}}\toprule
Arm & Recall [95\%] & False alarms/frame & GMAC/frame \\\midrule
Cue alone & 1.4\% & 0.16 & 0 \\
Dense network & 5.7\% [3.1, 9.6] & 0.17 & 1.43 \\
Gated, cap 128 & 10.4\% [6.6, 15.1] & 0.19 & 0.28 \\
Gated + exploration (deployed) & 10.5\% & 0.20 & 0.43 \\
YOLO11n, same folds & 54.5\% [49.9, 58.9] & 0.14 & 3.2 \\\bottomrule
\end{tabular}
\end{table}

\subsection{Edge--ground split on mountain search imagery}\label{sec:split}
Table~\ref{tab:split} is the primary read-out: thresholds set on each fold's validation location for targets of 0.5 and 1.0 false alarms per frame, then applied unchanged to the test locations. We report recall and the false-detection rate actually reached at the unseen test locations. That rate counts detections per frame; the burden of distinct operator alerts is a separate quantity (Section~\ref{sec:open}).

\begin{table}[h]\centering\small
\caption{HERIDAL (Zenodo 5662351, deduplicated), 15 test locations, thresholds frozen on the validation location. Recall [95\%] and actual test false alarms per frame. YOLO rows are from the Y2 run, which retrained YOLO11n per fold; the 12-epoch control row is from the follow-up run on the same crops.}\label{tab:split}
\begin{tabular}{@{}lcc@{}}\toprule
Arm & Target 0.5 & Target 1.0 \\\midrule
Our board, all candidates & 50.9\% [45.9, 58.2] at 2.24 & 59.8\% [53.9, 65.5] at 3.61 \\
Our board, top 16, board scores (exploratory) & 47.5\% at 1.92 & 53.4\% at 2.79 \\
Our board, same 16, ground scores & 46.7\% [40.4, 61.3] at 0.80 & 51.7\% [45.1, 66.8] at 1.32 \\\midrule
YOLO11n, all detections & 45.1\% [38.0, 61.0] at 0.87 & 53.1\% [46.7, 67.0] at 1.68 \\
YOLO11n, top 16, YOLO scores & 44.9\% [37.7, 61.0] at 0.86 & 52.5\% [45.9, 67.0] at 1.63 \\
YOLO11n, same 16, ground scores & 53.7\% [46.2, 66.3] at 0.64 & 61.6\% [55.4, 74.4] at 1.15 \\
Same 16, shuffled-label, 12 epochs & 6.6\% [3.3, 13.2] at 1.04 & 9.4\% [5.7, 17.0] at 1.50 \\
Same 16, shuffled-label, 3 epochs & 2.8\% at 0.89 & 5.4\% at 1.34 \\\midrule
\multicolumn{3}{@{}l}{\emph{Paired change, ground scores minus YOLO scores, same 16 crops, 15 locations (post hoc)}} \\
Recall, points & +8.8 [+2.4, +12.5] & +9.2 [+4.7, +13.0] \\
False detections per frame & $-0.22$ [$-0.75$, +0.30] & $-0.48$ [$-1.38$, +0.35] \\\bottomrule
\end{tabular}
\end{table}

No detector run alone holds its false-alarm target at new locations. Our board overshoots by 3.6 to 4.5 times. YOLO11n overshoots by 1.7 times in this run, and by 3.1 to 3.4 times in an earlier run under the same recipe (52.6\% at 1.70 and 59.0\% at 3.08). With the ground model behind it, either detector's pooled rate stays within 1.6 times of the target; behind YOLO11n the paired change is not distinguishable from zero (below).

Each detector has three arms that isolate the ground model. For YOLO11n all three were declared before the run; for our board, the top-16 row was printed by the same scorer but not declared, and is exploratory. The arms are all detections; the top 16 with the detector's own scores; and the same 16 re-scored by the ground model. Every arm sets its threshold on validation and freezes it. Capping YOLO11n at 16 candidates changes almost nothing (45.1\% at 0.87 against 44.9\% at 0.86), so the change from the cap to the ground arm is due to the ground scores. Against the same 16 crops, the ground model raises pooled recall by 8.8 and 9.1 points, and pooled false detections fall from 0.86 to 0.64 and from 1.63 to 1.15 per frame. Behind our board, the same comparison shows no recall gain (47.5\% to 46.7\%) and false detections cut 2.4 and 2.1 times.

The separate intervals overlap, so we estimated the paired changes post hoc: the same 15 test locations for both arms, each arm with its own frozen threshold, pooled recall and false detections per frame, and 2,000 bootstrap draws of locations applied to both arms together. Recall rose by 8.8 points [+2.4, +12.5] at the 0.5 target and 9.2 points [+4.7, +13.0] at the 1.0 target; it rose at 10 and 11 of the 15 locations. The change in false detections per frame was $-0.22$ [$-0.75$, +0.30] and $-0.48$ [$-1.38$, +0.35]; both intervals include zero, and false detections fell at only 6 and 5 locations, with large drops at four locations and rises at five. One location holds 1,312 of the 3,091 test people and contributes about half the recall gain; resampling by location accounts for this. We therefore claim higher recall over the same candidates in this run, and no reduction in false detections.

A shuffled-label control trained for the same 12 epochs on the same crops reaches 6.6\% and 9.4\% recall; its training loss stays flat from the first epoch, and the real ground model is above it at every location. Training budget does not explain the gap. Sending 16 crops costs coverage: YOLO's top 16 contains the person in 71\% of cases against 81\% for all its detections, a loss of 10 points, or about 12\% of what unrestricted YOLO covers.

On the test-set curves, a secondary read-out chosen after the first run, the split behind our board gives 41.0\% and 48.7\% at 0.5 and 1.0 false alarms per frame against 24.1\% and 34.9\% for the board alone; behind YOLO11n it gives 49.7\% and 59.7\% against 36.9\% and 46.8\%.

\subsection{Power on the board}
LaCASA at the University of Alabama in Huntsville ran our pipeline on a Raspberry~Pi~5 (8~GB, CPU, active cooler) at two frames per second, three runs per mode, with a Joulescope JS220 at 2~kHz~\cite{dzeletovic2026}. We aligned each trace to the software's markers and accepted a run if the alignment correlation reached 0.8, a gate set before the traces arrived. The powersave-governor runs passed (0.895 to 0.900). Runs at the default governor and at full speed did not, and we report no energy from them. An outside reviewer recalculated all nine runs and matched our figures to the third decimal.

\begin{table}[h]\centering\small
\caption{Raspberry Pi 5, powersave governor, 2 frames per second, mean of three runs [range]. Idle board 2.28~W. The bench frames keep the cue at its 128-place cap, so this is a heavy-load case. YOLO11n was not measured on the board.}\label{tab:power}
\begin{tabular}{@{}lccc@{}}\toprule
Mode & Above idle (W) & Whole board per frame (J) & Frames/s at full speed \\\midrule
Cue only & 0.05 & 1.16 & 50.7 \\
Dense network & 0.43 [0.40, 0.50] & 1.36 & 15.0 \\
Gated & 0.28 [0.25, 0.30] & 1.28 & 16.4 \\\bottomrule
\end{tabular}
\end{table}

Gating cuts power above idle by 32 to 39\%. Because idle dominates, the whole-board saving is about 6\%. We make no power claim for YOLO11n: it was not run on the Pi, and arithmetic alone did not predict power for our own arms.

\subsection{Field check}\label{sec:field}
Table~\ref{tab:field} gives candidate coverage on our footage: the share of samples where the person is among the top 16 points. Our board detector and YOLO11n trained on HERIDAL saw neither this site nor this camera.

\begin{table}[h]\centering\small
\caption{Walking person, our footage, person among the top 16. Our board is the mean over three folds. YOLO11n trained on HERIDAL; G is the mean over the same three folds. Passes are 640-equivalent YOLO inferences per frame.}\label{tab:field}
\begin{tabular}{@{}lcccccc@{}}\toprule
Height ($n$) & Our board & W640 & W1280 & Tiles T & Board-chosen G & COCO, best \\\midrule
117.5~m (109) & 74\% & 1.8\% & 89.0\% & 91.7\% & 79.8\% & 15.6\% \\
89.2~m (8) & 50\% & 37.5\% & 50.0\% & 87.5\% & 66.7\% & 12.5\% \\
60.3~m (26) & 40\% & 23.1\% & 46.2\% & 50.0\% & 53.8\% & 17.9\% \\
9.3~m (52) & 92\% & 69.2\% & 17.3\% & 3.8\% & 28.2\% & 88.5\% \\\midrule
Passes per frame & n/a & 1 & 4 & 24 & 9--11 & \\\bottomrule
\end{tabular}
\end{table}

\begin{figure}[t]
\centering
\includegraphics[width=\linewidth]{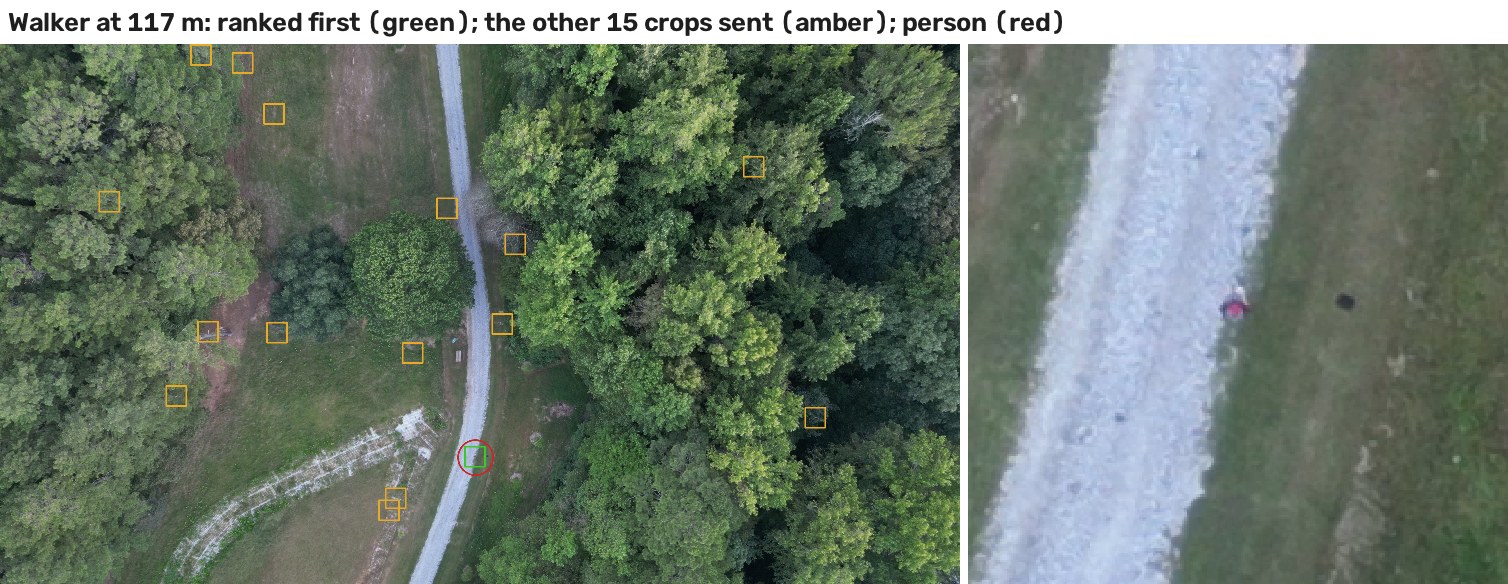}
\caption{One frame of our footage at 117~m. Our board ranked the walker first (green); amber boxes are the other 15 crops sent to the ground; the red circle is the labelled person. Right: the walker enlarged.}\label{fig:field}
\end{figure}

At 117.5~m and 89.2~m, YOLO11n trained on HERIDAL has the walker in its top 16 more often than our board, at 4 to 24 passes per frame. Windows chosen by our board do not beat YOLO's own tiling at the two upper heights, so the idea that our board can choose where YOLO looks is not supported on this footage. The HERIDAL-trained configurations that do best at altitude do worst at 9.3~m, where the COCO weights do best; off the shelf, the COCO weights find the walker at most 18\% of the time above 60~m. The walker wore red on bright gravel (Figure~\ref{fig:field}).

For the still person in regular clothing, lying in the open at 117~m, sitting at a tree at 90~m and half under canopy at 60~m, our board had the person in its top 16 in 0 to 1\% of samples ($n$ = 51, 44, 64), and the best YOLO11n configuration in 0 to 7.5\%. With a bright vest, YOLO11n did much better at 90 and 60~m (up to 100\% and 95\%, against 40\% and 20\% for our board) and both were weak at 117~m (20\% and 33\%). Groups have 5 to 83 samples and no interval was computed. Coverage of a still person in earth tones was therefore very low in the tested clips. The experiment does not separate the possible causes: altitude, clothing, pose, occlusion, image resolution and training data change together across these clips. Thermal imaging has found hidden people where colour did not~\cite{schedl2021}, which makes it a candidate worth testing here.

\section{Considerations for deploying a standard detector on the board}\label{sec:open}
A standard detector is the stronger choice on accuracy, on thermal imagery and at altitude. We did not run YOLO11n on the Pi, so its timing and power there are unknown. Considering it for board deployment, our measurements raise four issues, and two more remain to be measured.

\begin{enumerate}
\item \emph{False alarms at new places.} Thresholds set at one location overshot their target 1.7 to 3.4 times at new locations. A threshold calibrated to one false detection per frame produced up to 3.08 false detections per frame at the test locations. Behind the ground model the pooled overshoot was at most 1.6 times, but the paired change in false detections behind YOLO11n was not distinguishable from zero.
\item \emph{Cost of seeing small people.} At 117~m the best YOLO11n configurations need 4 (whole frame at 1280) to 24 (tiles) standard passes per 4K frame, about 13 to 77~GMAC, against 3.2 for one pass and 0.28 for our gated arm. On a desktop x86 CPU one pass took about 15~ms with two threads; we have no Pi timing.
\item \emph{No one setting for all heights.} The configuration that did best at 117~m found 3.8\% at 9.3~m; the one that did best at 9.3~m (COCO weights, board-chosen windows) found 10.7\% at 117~m. These results motivate testing an altitude-aware input policy on independently held-out flights.
\item \emph{Still people in earth tones.} Coverage was 0 to 7.5\% for both detectors from 60~m up in our clips. We do not yet know which of altitude, clothing, pose, occlusion, resolution or training data is responsible.
\item \emph{Power and heat on the board} (not measured). Our own arms showed that arithmetic does not predict board power; YOLO11n's 4 to 24 passes per frame must be measured on the Pi, with and without an accelerator.
\item \emph{Licence} (a fact, not a measurement). Ultralytics YOLO is AGPL-3.0, which matters for closed deployments; a permissively licensed detector or a commercial licence would be needed.
\end{enumerate}

Against these, ground verification has costs. Table~\ref{tab:ops} turns the measured quantities into operational ones; these are calculations under the stated assumptions, not further measurements.

\begin{table}[h]\centering\small
\caption{Operational reading of the ground-verification arm behind YOLO11n at 2 processed frames per second.}\label{tab:ops}
\begin{tabularx}{\linewidth}{@{}lX@{}}\toprule
Measured & Implies \\\midrule
154~KB of crops per frame & About 2.46~Mbit/s per drone before protocol overhead; about 9.9~Mbit/s for four drones. \\
0.64 false detections per frame & About 77 frame-level false detections per minute per drone. One background object can repeat across frames, so this is not a count of distinct operator alerts. \\
16 crops at 0.43~GMAC & About 6.9~GMAC per frame on the ground computer. \\
Top 16 instead of all detections & 10 points less candidate coverage (81\% to 71\%). \\\bottomrule
\end{tabularx}
\end{table}

The crop payload is a bandwidth cost. Whether it is a bandwidth saving depends on the alternative, such as compressed full frames, which we have not measured. The false-detection rate needs consolidation across frames and an event-level measure, false alerts per minute, before it says what an operator would see.

\section{Roadmap}
Each step below is a test we will declare before running, with the same folds and rules.
\begin{enumerate}
\item \emph{Measure YOLO11n on the board.} Latency, energy per frame and temperature on a Raspberry~Pi~5, CPU-only and with a neural accelerator, for whole-frame and tiled inputs, on the LaCASA bench with the same sync gate.
\item \emph{Altitude-aware input.} Use flight altitude from telemetry to choose input size or tiling per frame, and test the rule on held-out footage at heights it was not set on.
\item \emph{Confirm the ground-verification gain.} Repeat Y2 with the paired analysis declared in advance, in one pinned environment and with a second seed, and test site calibration from frames known to be empty, since ground scoring did not reliably lower false detections.
\item \emph{Build on airborne optical sectioning.} Use the fast, wide colour pass with ground verification to flag areas, and send a slow, low thermal pass over them using AOS~\cite{schedl2020,schedl2021}, with the permission of its authors for any use of their code. Repeat the still-person clips there, varying altitude, clothing, pose and occlusion one at a time.
\item \emph{The complete path.} Measure detection, crop encoding, radio transfer, ground inference, consolidation across frames and delivery of a distinct alert, with false alerts per minute as the operator measure, against a compressed-frame alternative on the same radio.
\item \emph{More field data and licence.} More people, clothing, terrain and seasons at a second site, with intervals per group, and the baseline repeated with a permissively licensed detector. Our own 46,000-parameter detector is a candidate low-power fallback, pending the power comparison in step 1.
\end{enumerate}

\section{Discussion}
Five findings carry over beyond this pipeline. First, evaluation protocol matters. On HIT-UAV every inferred recording block appears on both sides of the published split, and holding out blocks gave a recall estimate less than half as large at the same false-alarm target. That comparison is exploratory and does not isolate leakage from the other differences between the protocols, but the two protocols produced substantially different estimates. Anyone choosing a detector for search from benchmark numbers should ask how the split was made.

Second, our small gated detector beats the same network applied everywhere, and a standard detector beats both by a wide margin. A cue that removes background lets a small network run at a lower threshold for the same false-alarm budget, and it cuts arithmetic; on the board that became about a third less power above idle and about 6\% at the wall, because the idle board dominates.

Third, threshold transfer is the problem a search operator meets first, and a better detector alone did not solve it. Both detectors overshot their false-alarm budget at new locations. A ground model re-scoring the same 16 crops raised recall behind the standard detector, with the paired change clear of zero; its effect on false detections was mixed across locations, so it is not a fix for threshold transfer on this evidence.

Fourth, benchmark and field results have to be read together. The open problems Manzini and Murphy drew from a real search (realistic data, varied imagery, agreed measures) are the ones we tried to address in a small way: public benchmarks evaluated under held-out blocks and locations with frozen thresholds, and our own footage at four heights from a site and camera no model had seen, scored with a rule set before scoring. Neither half is sufficient alone. The field footage is small, and the benchmarks are not searches.

Finally, the field result we would most want a search planner to read is the still person. Both detectors rarely had a still person in earth tones among their top 16 at 60~m or higher, while the same person in a bright vest was often there. Benchmark recall alone does not tell a search team whether a drone will find the person they are looking for.

\section{Limitations}
The field footage is one site, two afternoons and one or two people, with truth derived from motion or registration, and its groups are small. HERIDAL results are single stills, so memory across frames is not in them. The Y2 run used newer software (PyTorch 2.15, Ultralytics 8.4.213) on a different machine than the earlier runs; its YOLO rows are internally consistent but should not be mixed with the earlier YOLO run, which we quote separately. The paired analysis of the YOLO11n gain is post hoc, from one run in one software environment; it has not been repeated. Thermal recording blocks are inferred from frame identifiers. The power figures are for our pipeline only, a heavy-load case at one governor setting, on a bench, not in flight. We make no claim about the probability of finding a person in a search, which depends on search planning and terrain as much as on the detector.

\section{Reproducibility}
The primary evaluations were written down with their read-outs before they ran, in a dated protocol file kept with the code; analyses added after a result are labelled post hoc there and in this paper, and deviations are recorded before results are read. Each run's software versions are recorded with its result files. Each number in this paper comes from a result file whose SHA-256 digest is recorded. Code, models, truth registers and result files are at \url{https://github.com/numberz-ai/aerial-person-detector}; YOLO11n weights are not redistributed. Our field footage is not released.

\section*{Acknowledgements}
This work was funded by Numberz.ai Inc. and received no government funding. We thank Vojin Dzeletovi\'c and Dr.~Aleksandar Milenkovi\'c of the LaCASA Laboratory at the University of Alabama in Huntsville for the power measurements, Hunter Klobucar, an FAA Part~107 certificated remote pilot, who planned and flew the field flights and recorded the video and flight logs used in this paper, and Praveen Mohan for running the standard-detector tests. We thank the authors of HIT-UAV, HERIDAL and the small-person dataset of Makrigiorgis et al.\ for releasing their data.

\appendix
\section{Training details}\label{app:train}
\paragraph{Board network.} Input 44$\times$44 (one channel thermal, three colour), convolutions 8, 8, 16, 16, 32 and 32 channels with batch normalization and ReLU, two 2$\times$2 max-pools, and a 1$\times$1 output; 45,945 parameters on one channel and 46,089 on three. Binary cross-entropy, AdamW (learning rate $2\times10^{-3}$, weight decay $10^{-4}$) with a one-cycle schedule, batch 256, 8 epochs. Positives are labelled people; negatives, three per positive, are drawn from cue candidates, random points and vehicle centres. Augmentation: shifts of up to 2~px, 90-degree rotations, flips, gain and offset changes and Gaussian noise. Thermal training uses image scales 1.0, 0.75, 0.5, 0.35 and 0.25; HERIDAL images are processed at scale 0.2 (800$\times$600) with training scales 1.0, 0.75 and 0.5 relative to that.

\paragraph{Thermal cue grid.} $z \in \{1.5, 2.0, 2.5\}$ and $\delta \in \{4, 6, 8\}$, chosen per fold by cue ceiling on 200 frames from that fold's training and validation blocks at scales 1.0, 0.5 and 0.25; ties within 0.5 points go to fewer candidates, then to the earlier default (2.0, 6). Folds 0 to 3 chose (1.5, 8) and fold 4 the default.

\paragraph{Ground confirmer.} Four blocks of two 3$\times$3 convolutions (32, 64, 128 and 192 channels) with batch normalization, ReLU and max-pooling, then global average pooling, dropout 0.2 and one linear output; 842,209 parameters. Trained per fold from scratch with binary cross-entropy and AdamW on the detector's top 32 points per training frame, labelled hit or miss, plus a jittered crop on each person; 12 epochs. Shuffled-label controls: 3 epochs behind our board and in Y2; 12 epochs, same permutation and seeds, in the Y2 follow-up.

\paragraph{Thresholds.} For each budget, the threshold is the lowest score whose false alarms per frame on validation do not exceed the budget.

\bibliographystyle{plainnat}
\bibliography{paper_refs}
\end{document}